# Neural shrinkage for wavelet-based SAR despeckling

Mario Mastriani, and Alberto E. Giraldez

*Abstract*—The wavelet shrinkage denoising approach is able to maintain local regularity of a signal while suppressing noise. However, the conventional wavelet shrinkage based methods are not time-scale adaptive to track the local time-scale variation. In this paper, a new type of Neural Shrinkage (NS) is presented with a new class of shrinkage architecture for speckle reduction in Synthetic Aperture Radar (SAR) images. The numerical results indicate that the new method outperforms the standard filters, the standard wavelet shrinkage despeckling method, and previous NS.

## I. INTRODUCTION

DESPECKLING a given speckle corrupted image is a traditional problem in both biomedical and in synthetic aperture processing applications, including synthetic aperture radar (SAR). In a SAR image, speckle manifests itself in the form of a random pixel-to-pixel variation with statistical properties similar to those of thermal noise. Due to its granular appearance in an image, speckle noise makes it very difficult to visually and automatically interpret SAR data. Therefore, speckle filtering is a critical preprocessing step for many SAR image processing tasks [1]-[6], such as segmentation and classification.

Many algorithms have been developed to suppress speckle noise in order to facilitate postprocessing tasks. Two types of approaches are traditionally used. The first, often referred to as multilook processing, involves the incoherent averaging of *L* multiple looks during the generation of the SAR image. The averaging process narrows down the probability density function (pdf) of speckle and reduces the variance by a factor *L*, but this is achieved at the expense of the spatial resolution (the pixel area is increased by a factor). If the looks are not independent, such as when the Doppler bandwidth of the SAR return signal is segmented into multiple overlapping subbands, one needs to define an equivalent number of looks (ENL) [7] to describe the speckle in the resultant images. The second approach, which is applied after the formation of the multi-look SAR image, involves the use of adaptive spatial filtering through an examination of the local statistics surrounding a given pixel [2], [3]. To date, various spatial filters have been developed to reduce speckle without significant loss in spatial resolution. The best known filters include those by Lee [8], Kuan [9], Frost [10], their own variations such as the enhanced Lee filter [11], the refined Lee filter [12], the enhanced Frost filter [11], and many others (see [13]–[15]). A good adaptive speckle filter should possess the following properties [7]: speckle reduction in statistically homogeneous areas; feature preservation (such as edges and real textural variations); radiometric preservation.

A spatial filter's performance depends heavily on the choice of the local window size and orientation. As stated in [16], and also noted by other observers, "the spatial organization of a surface's reflectance function is often generated by a number of different processes, each operating at a different scale." As a result, features present in SAR imagery often exhibit different scales. This requires an adjustable window to adapt to local spatial variations, including the feature scale and geometric structure. Most filters fail to achieve spatial adaptation because they only deploy a local window with fixed size and shape. There exist a few filters that are capable of adapting the size or the shape of the local window according to the underlying structural features. The refined Lee filter [12] is such an example.

Wavelet multiresolution analysis has the very useful property of space and scale localization, so it provides great promise for image feature detection at different scales. In view of the many theoretical developments that occurred in the last decade, wavelets have found successful applications in a variety of signal processing problems, including image coding and image denoising. Recently, Donoho et al. [17]-[19] developed a nonlinear wavelet shrinkage denoising method for statistical applications. The wavelet shrinkage methods rely on the basic idea that the energy of a signal (with some smoothness) will often be concentrated in a few coefficients in wavelet domain while the energy of noise is spread among all coefficients in wavelet domain. Therefore, the nonlinear shrinkage function in wavelet domain will tend to keep a few larger coefficients representing the signal while the noise coefficients will tend to reduce to zero. On the other hand, recent wavelet thresholding based denoising methods proved promising [17], [20]-[22], since they are capable of suppressing noise while maintaining the high frequency signal details. However, the local space-scale information of the image is not adaptively considered by standard wavelet thresholding methods. In standard wavelet thresholding based noise reduction methods [21], [22], the threshold at certain scale is a constant for all wavelet coefficients at this scale.

A major difficulty in achieving adaptive algorithm using wavelet thresholding methods is that the soft-thresholding function is a piece-wise function and does not have any high order derivatives. Therefore, a new type of smooth nonlinear shrinkage functions is necessary [23] and a new class of NS results in consequence. Unlike the standard shrinkage functions, these new nonlinear shrinkage functions depend on speckle directly. Then a new nonlinear 2-D adaptive filtering method based on wavelet NS is presented for space-scale adaptive speckle reduction in SAR images.

## II. SPECKLE MODEL

Speckle noise in SAR images is usually modeled as a purely multiplicative noise process of the form

$$I_s(r,c) = I(r,c).S(r,c)$$
$$= I(r,c).[\,1+T(r,c)\,]$$
$$= I(r,c) + N(r,c) \quad (1)$$

The true radiometric values of the image are represented by $I$, and the values measured by the radar instrument are represented by $I_s$. The speckle noise is represented by $S$, and it modeled as a stationary random process independent of $I$, with $E[S] = 1$, where $E[\bullet]$ is the expectation operator of $[\bullet]$. The random process $T$ is zero mean, with variance $\sigma_T^2$ and known autocorrelation function $R_{TT} = R_{SS} - 1$. The parameters $r$ and $c$ means row and column of the respective pixel of the image. If

$$T(r,c) = S(r,c) - 1 \quad (2)$$

and

$$N(r,c) = I(r,c).T(r,c) \quad (3)$$

we begin with a multiplicative speckle $S$ and finish with an additive speckle $N$ [24], which avoid the log-transform, because the mean of log-transformed speckle noise does not equal to zero [25] and thus requires correction to avoid extra distortion in the restored image. Eq.(3) represents and additive zero-mean *image-dependent* noise term, which is proportional to the image to be estimated. Since $I$ is nonstationary in general, the noise $N$ will be nonstationary as well.

For single-look SAR images, $S$ is Rayleigh distributed (for amplitude images) or negative exponentially distributed (for intensity images) with a mean of $1$. For multi-look SAR images with independent looks, $S$ has a gamma distribution with a mean of $1$. Further details on this noise model are in [26].

## III. WAVELET-BASED DESPECKLING

The discrete wavelet transform (DWT) [17]-[19], [27] of a one-dimensional (1-D) signal is implemented by two-channel subband filtering followed by downsampling by a factor of two. The two filters [3], [22] including $\{h(k)\}$, the scaling filter (lowpass), and $\{g(k)\}$, the wavelet filter (highpass), constitute a pair of quadrature mirror filter (QMF) banks [28]. The transformation in two dimensions can be readily derived in a straightforward manner from 1-D [17]-[19], [27]. At each level, the decomposition scheme applies the scaling filter and the wavelet filter alternately to the rows and columns of the two-dimensional (2-D) image under analysis [2], [3]. At any decomposition level $j = 1, \ldots, J$, the input is transformed into four subbands. By their frequency contents, they are named the approximation subband $LL_j$ and three detail subbands $LH_j$ ($L$ stands for lowpass filtering, and $H$ stands for highpass filtering), $HL_j$, and $HH_j$. Since the approximation subband $LL_j$ contains the low-frequency portion of the original image, it carries most of the original information, whereas the detail subbands $LH_j$, $HL_j$, and $HH_j$ capture the horizontal, vertical, and diagonal features in the image respectively. Subband $LL_j$ will be used as an input for further decomposition to obtain multiscale analysis at level $j+1$. At level 0, $LL_0$ is represented by the original image.

The essence of denoising using wavelet analysis is to reduce the noise in the wavelet transform domain. Suppose we have a length-$M$ noisy observation $I_s = [I_{s,1}, I_{s,2}, \ldots, I_{s,M}]$

$$I_s = I + N \quad (4)$$

where $I = [I_1, I_2, \ldots, I_M]$ is the desired noise-free signal, and $N = [N_1, N_2, \ldots, N_M]$ is the observation noise. Because a DWT is a linear operator, it yields an additive noise model in the transform domain

$$y = \text{DWT}(I_s)$$
$$= \text{DWT}(I) + \text{DWT}(N)$$
$$= x + n \quad (5)$$

If $n$ is an equivalent additive speckle model (EASM) [24] with zero mean and standard deviation $\sigma_n$, $n$ shall remain approximately white Gaussian with zero mean and standard deviation $\sigma_n$ [24] because of the orthonormal property of wavelet basis functions. In the wavelet despeckling problem, we assume the speckle noise to approximately follow a Gaussian distribution. According to the central limit theorem, the noise in the transform domain will approach Gaussianity more closely. In order to simplify notation, above we use the 1-D vector format with boldfaced letters to represent 2-D images instead of the matrix representation. For the $i$th wavelet coefficient at level $j$ in detail subband $d$ ($d = 1$, $HL$; $d = 2$, $LH$; $d = 3$, $HH$), the observation model in the wavelet domain is formulated more specifically by

$$y_{i,j}^d = x_{i,j}^d + n_{i,j}^d \qquad (6)$$

For clarity of notation, we will omit the level index $j$ and the detail subband index unless they are explicitly needed.

The main scheme for recovering $x$ from $y$ using the wavelet transform can be summarized by the three primary steps shown in the block diagram in Fig. 1, i.e.,

1) Calculate the bidimensional Discrete Wavelet Transform (DWT-2D) of the speckled image.
2) Modify the speckled wavelet coefficients according to some rule.
3) Compute the inverse of DWT-2D (IDWT-2D) using the modified coefficients.

In the vast majority of wavelet despeckling algorithms, speckle reduction is accomplished in the detail subbands with the approximation subband not subjected to any changes.

In general, manipulating the wavelet coefficients is the most crucial step. What distinguishes one denoising method from another is mainly related to the approach used in this particular step. Loosely speaking, two major denoising techniques used in this context are the thresholding technique and the Bayesian estimation shrinkage technique. In these two techniques, algorithms can be further categorized by how the wavelet coefficients are statistically modeled. Most early models [29], [31] assumed the wavelet coefficients to be independently distributed. As the wavelet transform deepened its application in image coding and denoising, researchers proposed more complicated but also more accurate models that exploit interscale dependencies [33], intrascale dependencies [32], [34], [35], and the hybrid inter- and intrascale dependencies [30], [36] among wavelet coefficients. We will discuss some algorithms briefly in Section III-A, III-B and III-C.

### A. Thresholding Technique

Denoising based on thresholding in the wavelet domain was initially proposed in [17] (see also [19]). Thresholding typically involves a binary decision. The corresponding manipulation of wavelet coefficients usually consists of either "keeping (shrinking)" or "killing" the value of the coefficient. In [19], the authors introduced two thresholding methods, namely soft and hard thresholding. For each wavelet coefficient, if its amplitude is smaller than a predefined threshold, it will be set to zero (kill); otherwise it will be kept unchanged (hard thresholding), or shrunk in the absolute value by an amount of the threshold (soft thresholding).

The key decision in the thresholding technique is the selection of an appropriate threshold. If this value is too small, the recovered image will remain noisy. On the other hand, if the value is too large, important image details will be smoothed out. Using a minimax criterion, Donoho [19] proposed what the wavelet community calls the universal threshold $T = \sqrt{2\log(M)}\,\sigma_n$, where $N$ is the sample size, and $\sigma_n$ is the noise standard deviation. The universal thresholding technique has been recognized as simple and efficient, but when only a single threshold is used globally, it provides no spatial adaptation during the process of noise suppression [17]-[19]. In addition, studies have shown that with a very large sample size, the universal threshold tends to smear out details. Following [19], some researchers have focused on developing spatially adaptive thresholding techniques instead of using a global uniform threshold. In [37], a simple scaling factor function was proposed to regulate thresholds for the purpose of scale adaptation. Chang [32] first proposed a multiple threshold denoising scheme to take into account local spatial characteristics. In that work, the image of interest is first segmented into three major categories: edges, textured areas, and homogeneous areas. Then, thresholding is carried out with three different thresholds adapted to the three spatial categories. The limit of that method arises from the fact that the three different thresholds are selected in an *ad hoc* way. Using the Gaussian distribution and Laplacian distribution to model wavelet coefficients, Chang *et al.* [31] proposed an approximate minimum mean-square error (MMSE) solution to soft-thresholding. The so-called BayesShrink threshold is calculated as $T = \sigma_n^2 / \sigma_x$, where $\sigma_n^2$ and $\sigma_x^2$ are the noise variance and the image variance, respectively. This threshold is designed to adapt to each individual subband at each resolution level.

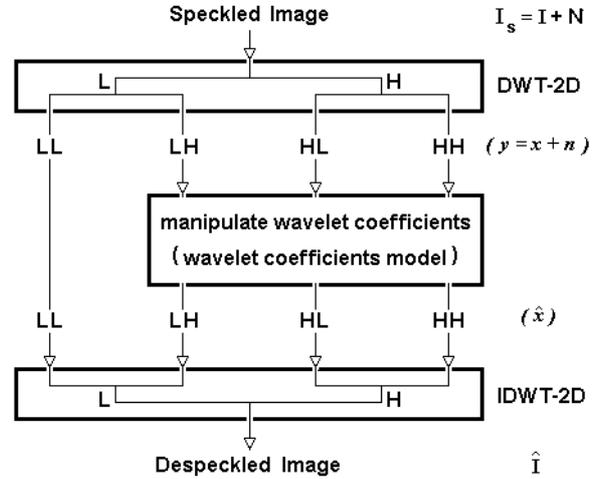

Fig. 1: The wavelet despeckling procedure with equivalent additive speckle model (EASM).

### B. Bayesian Estimator

As far as Bayesian estimation is concerned, it is necessary to assume an *a priori* distribution $p(x)$ associated with the wavelet coefficients of the noise-free image. If we know the likelihood function $p(x|y)$, we can estimate the noise-free wavelet coefficients $x$ by either of the following approaches [36]:

- Maximum *a posteriori* (MAP) estimator:

$$\hat{x} = \arg\max p(x|y)$$
$$= \arg\max p(y|x)p(x) \quad (7)$$

- MMSE estimator:

$$\hat{x} = E(x|y) = \int x\, p(x|y)\, dv \quad (8)$$

In general, the Bayesian solution will end up with a continuous shrinking function imposed on noisy observations, in contrast with the thresholding method, which usually involves a binary thresholding action.

If we assume that is independently and identically Gaussian distributed with zero mean and variance $\sigma_x^2$, given the EASM, the MAP and MMSE estimators provide the same solution

$$\hat{x}_i = \frac{\sigma_x^2}{\sigma_x^2 + \sigma_n^2} y_i \quad (9)$$

The deficiencies associated with this shrinking function are twofold. First, the assumed prior disagrees with the strong non-Gaussian statistics exhibited by wavelet coefficients of natural images. Secondly, each wavelet coefficient is denoised individually with the lack of spatial adaptation toward the intrascale and interscale dependencies.

For a Bayesian estimation process to be successful, the correct choice of priors for wavelet coefficients is certainly a very important factor. Several different priors have been considered for the wavelet coefficients. In [35], wavelet coefficients are modelled as conditionally independent Gaussian random variables with locally adaptive variance, and then the MMSE solution is derived to estimate the noise-free wavelet coefficients. Many studies support the fact that the generalized Gaussian distribution (GGD) provides a good fit to the statistics of natural images [38], [39]. Unfortunately, in the Bayesian estimation process, there usually does not exist a closed-form solution for the estimate of noise-free wavelet coefficients when the signal prior is described by the GGD [39]. In most cases, numerical approaches have to be applied to obtain the solution. In order to cast the estimation problem into a mathematically friendly environment, a mixture density model has been recently proposed as a prior to statistically model the wavelet coefficients [29], [28]. Applying a mixture of two Gaussian distributions with one mixture component corresponding to insignificant coefficients (representing "homogeneity"), and the other to significant coefficients (representing "heterogeneity"), Chipman [29] reconstructed the noise-free signal as a nonlinear rescaling of noisy measurements using a simple but elegant closed-form representation.

A close examination shows that Chipman's algorithm accurately models the wavelet coefficients, but it fails to incorporate the spatial dependence between wavelet coefficients into the denoising procedure. From visual inspection, we find that important wavelet coefficients tend to cluster at the location where signal transitions occur in the image domain. In [33], a hidden Markov tree model was proposed to capture the interscale dependence of wavelet coefficients. By measuring mutual information, Liu and Moulin [40] have shown that "intrascale models capture most of the dependencies between wavelet coefficients, and the gains obtained by including interscale dependence are marginal." In this study, we propose therefore to adopt the mixture of Gaussian densities to model natural images, meanwhile characterizing the intrascale contextual dependence of wavelet coefficients using Markov random fields (MRF). By properly fusing Bayesian estimation and Markov random field modeling, our goal is to achieve spatially adaptive wavelet despeckling. The idea of exploiting the clustering property of wavelet coefficients using MRF also appears in [34]; however, the approach differs from our proposed study mainly in its assumed prior probability for wavelet coefficients and the resultant shrinking function. In [34], the prior probability function is assumed as a piecewise continuous potential function with two constant parts and a linear transition around a predefined threshold. The corresponding manipulation part is implemented by multiplying each coefficient with its marginal probability of being a significant coefficient given all the observation data.

### C. Thresholding Neural Networks

Zhang constructs a type of thresholding neural network (TNN) to perform the thresholding in the transform domain to achieve noise reduction [20], [21], [23], [41]. The neural network structure of the TNN is shown in Fig. 2.

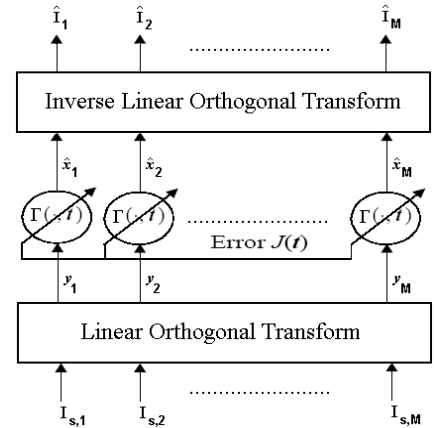

Fig. 2: Thresholding neural networks, after [23].

The transform in TNNs can be any linear orthogonal transform. The linear transform performed on observed data samples can change the energy distribution of signal and noise samples. By thresholding, the signal energy may be kept while the noise is suppressed. For a specific class of signal, the appropriate linear transforms may be selected to concentrate signal energy versus noise, and then a good mean-square-error (MSE) performance can be achieved. Here the thresholding functions are employed as nonlinear activation functions of the neural network. The inverse transform is employed to recover the signal from the noise-reduced coefficients in the transform domain. Specifically, since most signals have some kinds of regularities and the wavelet transform is a very good tool to efficiently represent such characteristics of the signal, the wavelet transform is often a suitable linear transform in TNNs. Note that there are several orthogonal channels in the transform domain. The input to the TNN is noise corrupted signal samples

$$I_{s,i} = I_i + N_i \qquad (10)$$

with $i = 0, ..., M-1$, where $I$ is the true signal and $N$ is the additive noise. The transform in TNNs can be any linear orthogonal transform. For a specific class of signal, the appropriate linear transform may be selected to concentrate signal energy versus noise. By thresholding, the signal energy may be kept while the noise is suppressed. Here the thresholding function $\Gamma(x,t)$ is employed as nonlinear activation functions of the neural network. The inverse transform is employed to recover the signal from the noise-reduced coefficients in transform domain, see Fig. 2. The different thresholds $t_j$ are used in different orthogonal channels and they are independent, i.e., the thresholds of different orthogonal channels can be optimized independently. It is also worth pointing out that although the term "neural network" is used, the TNN is different from the conventional multilayer neural network. In TNNs, a fixed linear transform is used and the nonlinear activation function is adaptive, while in conventional multilayer neural networks, the activation function is fixed and the weights of the linear connection of input signal samples are adaptive. We use the term "neural network" because the TNN has some basic elements similar to a conventional neural network, i.e., interconnection of input signal samples, nonlinear activation functions, and adaptivity to a specific input, etc. In addition, it is possible to change the fixed linear transform in Fig. 2 to an adaptive linear transform. In this way, both the weights of linear connections of input signal samples and the nonlinear activation function are adaptive, and then the conventional multilayer neural network techniques may be incorporated. This will be a meaningful exploration we are going to pursue in the future. However, the adaptive nonlinear activation function of Zhang's TNNs has an unacceptable computational cost, besides, it depends on the correct election of the initial activation function, on the other hand, the specific distributions of the signal and noise may not be well matched at different scales, etc. Therefore, a new method without these constraints will represent an upgrade.

## IV. NEURAL SHRINKAGE

Feed-forward neural network (FFNN) is the elected neural network for the NS implementation [42], which is directly applied to Coefficients of Detail (LH, HL and HH) of DWT-2D of Fig. 1. The neural network structure for our experiment can be illustrated as in Fig. 3.

Three layers, one input layer, one output layer and one hidden layer, are designed. The input layer and output layer are fully connected to the hidden layer. The training is achieved by designing the value of $K$, the number of neurones at the hidden layer, more than that of neurones at both input and the output layers. The training architecture is shown in Fig. 4, which shows a single level of decomposition for the approximation coefficients LL.

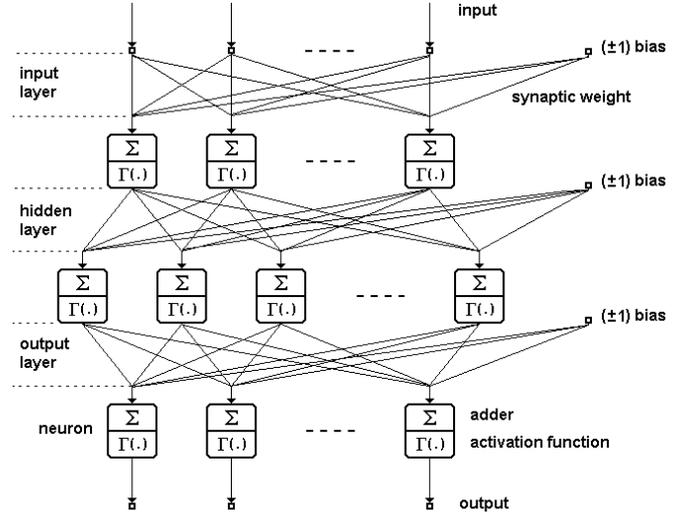

Fig. 3: Feed-forward neural network.

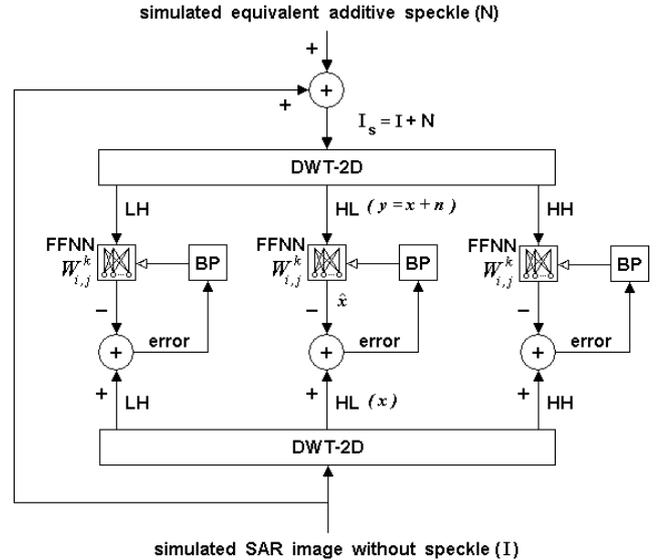

Fig. 4: Training architecture.

In Fig. 4, BP means back-propagation training algorithm [42], and the simulated speckle is summed to the original image, according to the Argenti et al's. approach [24].

The Fig. 5 represents the filter built with the NS, where, the trained neural networks represents to the optimal shrinkage function obtained in an automatic way. The method can be applied to each scale of decomposition of the Coefficients of Approximation, being much simpler that that methodology employed by Zhang [20], [21], [23], [41], and with a considerably smaller computational cost. While in Zhang [41], the activation function of the output layer is nonlinear and adaptive, here it is a linear function (the identity function), i.e., it is fixed, avoiding of this way the Zhang's problems. Finally, for hidden and input layers we use the hyperbolic tangent.

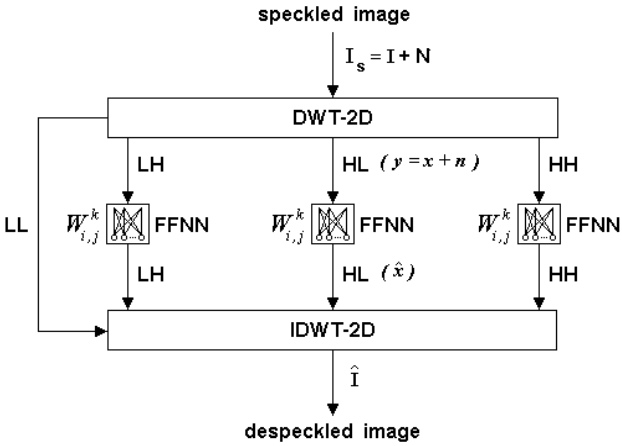

Fig. 5: Implementation of neural shrinkage.

### A. On the Optimal Performance of Neural Shrinkage

For any method dealing with the noise reduction problem, we want to get the bottom line of its performance. That is, we want to ask the questions:

- *What is the best performance of this method ?*
- *What are the properties of the optimal solution of the method ?*
- *How can the optimal solution of the method be achieved ?*

These are natural questions when we evaluate a method since the best performance defines all the potential of the method, and it is our objective to achieve the best performance of the method in practice. In this section, the optimal solution of the NS and its properties will be discussed. The learning algorithms of the NS to achieve the optimal solution will be presented in the next section.

Since the orthogonal linear transform used in the NS preserves the signal energy, the MSE of the estimation in the transform domain is equal to the MSE of the estimation in the time domain. Furthermore, since the thresholds of different orthogonal channels are independent and the thresholds of different orthogonal channels can be optimized independently, we will only analyze the optimal solution of one channel in the transform domain in the following, without loss of generality. In this context, and thanks to Argenti et al's approach [24] the speckle noise this doesn't imply a strategy change.

### B. Learning Algorithm for Neural Shrinkage

Let $y_i$ in Fig. 4 denote the space-scale data stream of the DWT-2D coefficients of the input speckled image $I_{s,i}$ and

$$y_i = x_i + n_i \qquad (11)$$

with $i = 0, ..., M-1$, in which $x_i$ is the space-scale data stream of the 2-D DWT coefficients of the true image and $n_i$ is the equivalent additive speckle noise in the transform domain. The objective is to obtain the estimate $\hat{x}_i$ of the true image DWT coefficients $x_i$, which minimize the MSE (Mean Square Error) risk

$$J(W) = E\{(x_i - \hat{x}_i)^2\} = \frac{1}{M}\sum_{i=1}^{M}\left[\Gamma(W_{i,j}^3\, y_i^3) - x_i\right]^2 \qquad (12)$$

In the new adaptive noise reduction scheme, the parameter $W_{i,j}^3$ is adaptively adjusted for the linear activation function $\Gamma(W_{i,j}^3\, y_i^3)$ to minimize the MSE, where $W_{i,j}^3$ denotes the synaptic weight matrix of output layer and $y_i^3$ is the output vector $[y_1^3, y_2^3, ...., y_M^3]^T$ of the hidden layer at the corresponding wavelet channel. In practice, since the original image $x$ is usually unknown, its DWT coefficients $x_i$ cannot be used as reference to estimate the risk $J(W)$.

Therefore, a practical reference is adopted: two noise corrupted signals $y$ and $y'$ are obtained from the same image $x$ plus uncorrelated noise $n$ and $n'$, and $y'$ is used as the reference. This is reasonable since in some applications, we may have an array of sensors and obtain more than one corrupted version of the signal. For example, in adaptive echo cancellation applications, two measurements for the same source signal are commonly used [43]. It can be proved that using such noisy reference signal leads to the same optimum threshold as using the true signal [20].

Usually a gradient-based LMS (Least Mean Square) stochastic adaptive learning algorithm [42], [43] is used for the NS to track local changes within the image and take advantage of the time-varying local estimation error, i.e., the synaptic weight matrix at layer *3* is adjusted by

$$\Delta W_{i,j}^3 = \mu\left(-\frac{\partial J}{\partial W_{i,j}^3}\right) \qquad (13)$$

with $i = 0, ..., M-1$, where the instantaneous error

$$\varepsilon_i = x_i - \hat{x}_i \qquad (14)$$

where $\mu$ is a learning parameter and $x_i$ is the space-scale data stream of the DWT-2D coefficients of the reference image y'. The synaptic weight matrices $W_{i,j}^k$ are dependent on not only different channels in transform domain but also spatial position, i.e., it is fully space-scale adaptive.

## V. ASSESSMENT PARAMETERS

In this work, the assessment parameters that are used to evaluate the performance of speckle reduction are
1) for simulated speckled images: Signal-to-Noise Ratio [36], and Pratt's figure of Merit [44].

2) for real speckled images: Noise Variance, Mean Square Difference, Noise Mean Value, Noise Standard Deviation, Equivalent Number of Looks, Deflection Ratio [2], [3], and Pratt's figure of Merit [44].

### A. Noise Mean Value (NMV), Noise Variance (NV), Mean Square Error (MSE), and Signal-to-Noise Ratio (SNR)

The SNR is defined as the ratio of the variance of the noise-free signal $I$ to the MSE between the noise-free signal and the denoised signal $\hat{I}$ [36]. The formulas for the NMV and NV calculation are

$$NMV = \frac{\sum_{r,c}\hat{I}(r,c)}{R*C} \quad (15)$$

$$NV = \frac{\sum_{r,c}(\hat{I}(r,c)-NMV)^2}{R*C} \quad (16)$$

where R-by-C pixels is the size of the despeckled image $\hat{I}$. On the other hand, the estimated noise variance is used to determine the amount of smoothing needed for each case for all filters. NV determines the contents of the speckle in the image. A lower variance gives a "cleaner" image as more speckle is reduced, although, it not necessarily depends on the intensity. The formulas for the MSE and SNR are

$$MSE = \frac{\sum_{r,c}(I(r,c)-\hat{I}(r,c))^2}{R*C} \quad (17)$$

$$SNR = 10\log_{10}\left(\frac{NV}{MSE}\right) \quad (18)$$

### B. Noise Standard Deviation (NSD)

The formula for the NSD calculation is

$$NSD = \sqrt{NV} \quad (19)$$

### C. Mean Square Difference (MSD)

MSD indicates average square difference of the pixels throughout the image between the original image (with speckle) $I_s$ and $\hat{I}$, see Fig. 7. A lower MSD indicates a smaller difference between the original (with speckle) and despeckled image. This means that there is a significant filter performance. Nevertheless, it is necessary to be very careful with the edges. The formula for the MSD calculation is

$$MSD = \frac{\sum_{r,c}(I_s(r,c)-\hat{I}(r,c))^2}{R*C} \quad (20)$$

### D. Equivalent Numbers of Looks (ENL)

Another good approach of estimating the speckle noise level in a SAR image is to measure the ENL over a uniform image region [3]. A larger value of ENL usually corresponds to a better quantitative performance.

The value of ENL also depends on the size of the tested region, theoretically a larger region will produces a higher ENL value than over a smaller region but it also tradeoff the accuracy of the readings. Due to the difficulty in identifying uniform areas in the image, we proposed to divide the image into smaller areas of 25x25 pixels, obtain the ENL for each of these smaller areas and finally take the average of these ENL values. The formula for the ENL calculation is

$$ENL = \frac{NMV^2}{NSD^2} \quad (21)$$

The significance of obtaining both MSD and ENL measurements in this work is to analyze the performance of the filter on the overall region as well as in smaller uniform regions.

### E. Deflection Ratio (DR)

A fourth performance estimator that we used in this work is the DR proposed by H. Guo et al (1994), [4]. The formula for the deflection calculation is

$$DR = \frac{1}{R*C}\sum_{r,c}\left(\frac{\hat{I}(r,c)-NMV}{NSD}\right) \quad (22)$$

The ratio DR should be higher at pixels with stronger reflector points and lower elsewhere. In H. Guo *et al*'s paper, this ratio is used to measure the performance between different wavelet shrinkage techniques. In this paper, we apply the ratio approach to all techniques after despeckling in the same way [2].

### F. Pratt's figure of merit (FOM)

To compare edge preservation performances of different speckle reduction schemes, we adopt the Pratt's figure of merit [44] defined by

$$FOM = \frac{1}{max\{\hat{N}, N_{ideal}\}} \sum_{i=1}^{\hat{N}} \frac{1}{1 + d_i^2 \alpha} \quad (23)$$

Where $\hat{N}$ and $N_{ideal}$ are the number of detected and ideal edge pixels, respectively, $d_i$ is the Euclidean distance between the $i$th detected edge pixel and the nearest ideal edge pixel, and $\alpha$ is a constant typically set to 1/9. *FOM* ranges between *0* and *1*, with unity for ideal edge detection.

## VI. EXPERIMENTAL RESULTS

### A. For Images with Simulated Speckle

Here, we present a set of experimental results using the NeuralShrink technique in standard 242-by-242 Lena image. The other methods against which we assess the performance of the proposed speckle filter include the following: the Bayesian soft thresholding technique proposed in [31] as described in Section III-A; the Bayesian MMSE estimation technique using the Gaussian mixture density model develop-ped in [29] and described in Section III-B; the refined Lee filter [12]; and the Wiener filter [23]. Fig. 6 shows the noisy image used in the experiment, and the filtered images. Table I shows the assessment parameters vs. 4 filters for Fig. 6.

Table I. Assessment Parameters vs. Filters for Fig. 6.

| Filters | Assessment Parameters | |
|---|---|---|
| | SNR | FOM |
| Noisy observation | 0.5432 | 0.37486 |
| Bayes soft thresholding | 0.8976 | 0.42311 |
| Bayes MMSE estimation | 0.8645 | 0.42387 |
| Refined Lee | 0.8712 | 0.42867 |
| Wiener | 0.8809 | 0.42111 |
| NeuralShrink | 0.9988 | 0.46121 |

### B. For Images with Real Speckle

Here, we present a set of experimental results using one ERS SAR Precision Image (PRI) standard of Buenos Aires area. For statistical filters employed along, i.e., Median, Lee, Kuan, Gamma-Map, Enhanced Lee, Frost, Enhanced Frost [2], [3], Wiener [23], DS [44] and Enhanced DS (EDS) [19], we use a homomorphic speckle reduction scheme [2], with 3-by-3, 5-by-5 and 7-by-7 kernel windows. Besides, for Lee, Enhanced Lee, Kuan, Gamma-Map, Frost and Enhanced Frost filters the damping factor is set to 1 [19], [3].

Fig. 7 shows a noisy image used in the experiment from remote sensing satellite ERS-2, with a 242-by-242 (pixels) by 65536 (gray levels); and the filtered images, processed by using VisuShrink (Hard-Thresholding), BayesShrink, Normal-Shrink, SUREShrink, and NeuralShrink techniques respectively, see Table II.

All the wavelet-based techniques used Daubechies 1 wavelet basis and 1 level of decomposition (improvements were not noticed with other basis of wavelets) [23], [31], [44]. Besides, Fig. 7 summarizes the edge preservation performance of the NeuralShrink technique vs. the rest of the shrinkage techniques with a considerably acceptable computational complexity.

Table II shows the assessment parameters vs. 19 filters for Fig. 7, where En-Lee means Enhanced Lee Filter, En-Frost means Enhanced Frost Filter, Non-log SWT means Non-logarithmic Stationary Wavelet Transform Shrinkage [24], Non-log DWT means Non-logarithmic DWT Shrinkage [25], VisuShrink (HT) means Hard-Thresholding, (ST) means Soft-Thresholding, and (SST) means Semi-ST [3], [31], [17]-[23]. We compute and compare the NMV and NSD over six different homogeneous regions in our SAR image, before and after filtering, for all filters. The NeuralShrink has obtained the best mean preservation and variance reduction, as shown in Table II. Since a successful speckle reducing filter will not significantly affect the mean intensity within a homogeneous region, NeuralShrink demonstrated to be the best in this sense too. The quantitative results of Table II show that the NeuralShrink technique can eliminate speckle without distorting useful image information and without destroying the important image edges. In fact, the NeuralShrink outperformed the conventional and no conventional speckle reducing filters in terms of edge preservation measured by Pratt's figure of merit [44], as shown in Table II. On the other hand, all filters were implemented in MATLAB® (Mathworks, Natick, MA) on a PC with an Athlon (2.4 GHz) processor.

## VII. CONCLUSION

In this paper, we developed a new type of NS structure for adaptive speckle reduction, which combines the linear filtering and shrinkage methods. We created new type architecture to serve as the shrinkage function of DWT-2D. Unlike the standard neural shrinkage techniques, the novel has a lower computational cost. By using this new shrinkage function, some gradient-based learning algorithms become possible and the learning process becomes more effective.

We then discussed the optimal solution of the NS in the MSE sense. It is proved that there is at most one optimal solution for the shrinkage representation of NS. The general optimal performance of NS is analyzed and compared to the linear speckle reduction method. It is shown that the shrinkage speckle reduction methods are more effective than linear methods when the signal energy concentrates on few coefficients in the transform domain. Besides, considerably increased deflection ratio strongly indicates improvement in detection performance. Finally, the method is computationally efficient and can significantly reduce the speckle while preserving the resolution of the original image, and avoiding several levels of decomposition and block effect.

ACKNOWLEDGMENT

M. Mastriani thanks Prof. Silvano Zanutto, Director of the Biomedical Engineering Institute, at University of Buenos Aires for his help and support.

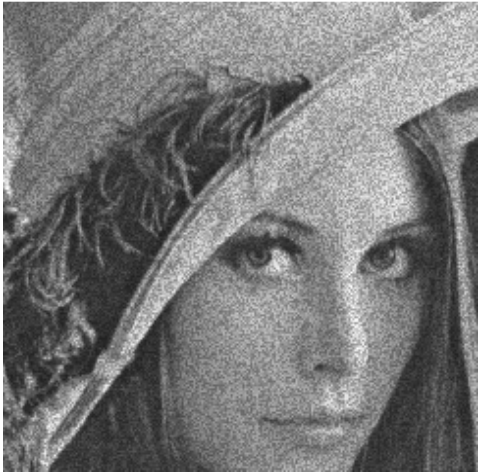

(a) Noisy observation

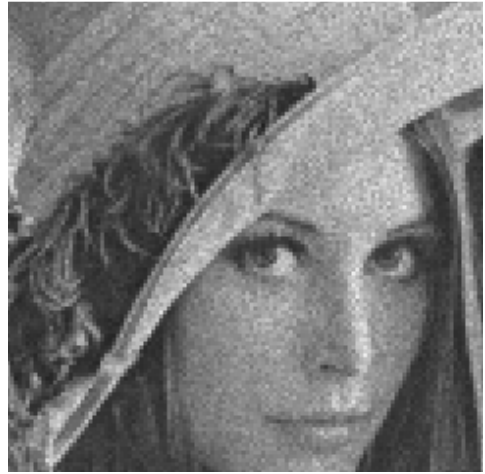

(b) Bayesian soft thresholding

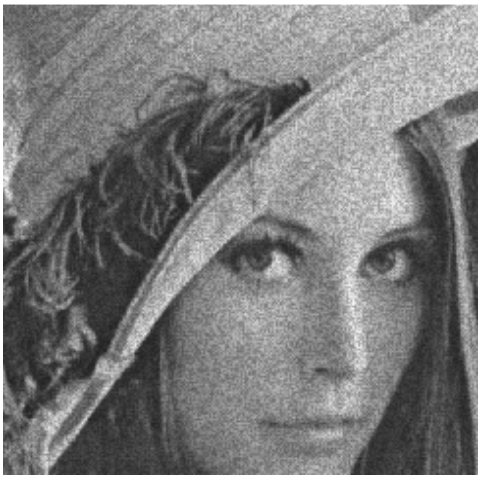

(c) Bayesian MMSE estimation

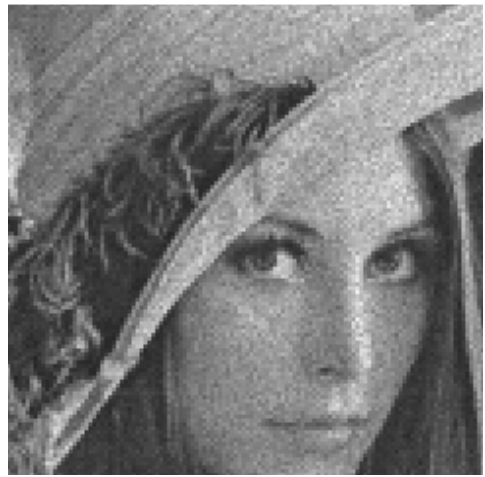

(d) Refined Lee

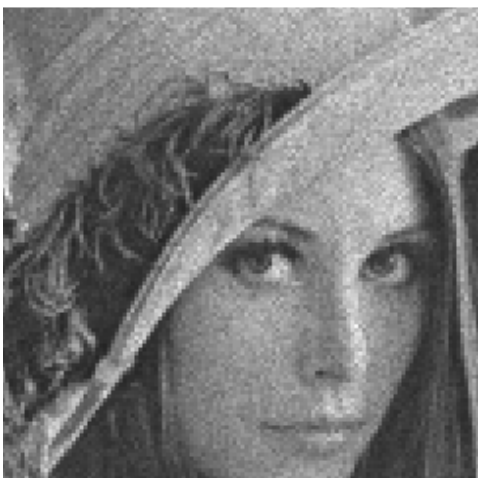

(e) Wiener

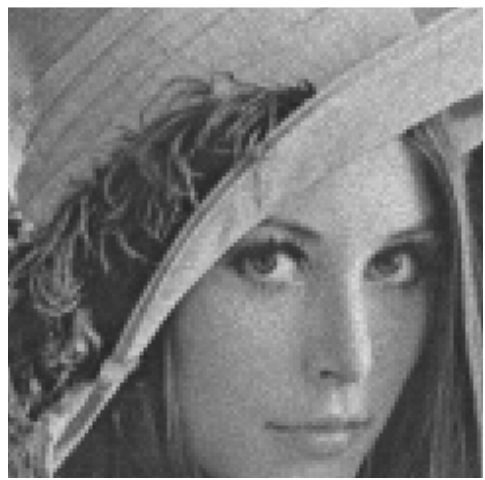

(f) NeuralShrink

Fig. 6: Lena with simulated noise and filtered.

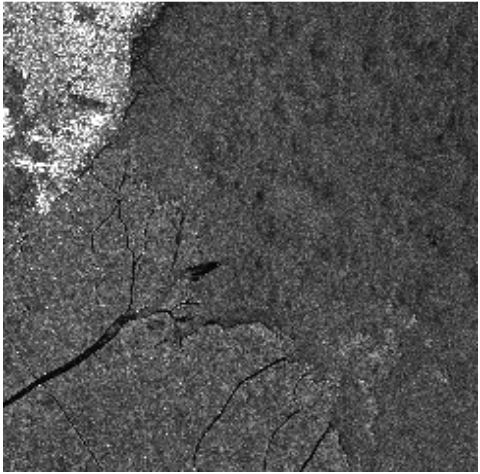

(a) Speckled observation

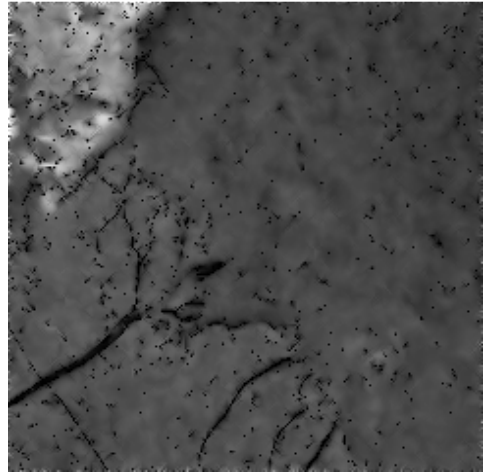

(b) Bayesian soft thresholding

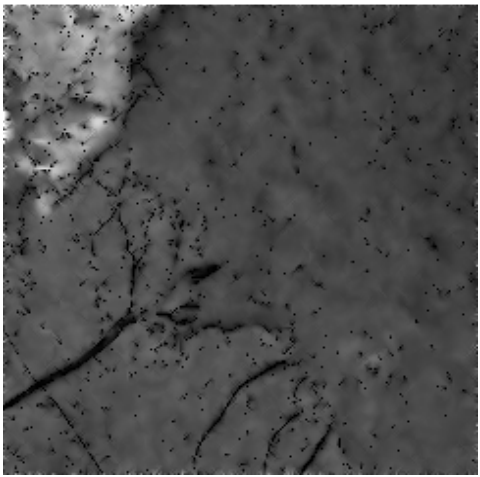

(c) Bayesian MMSE estimation

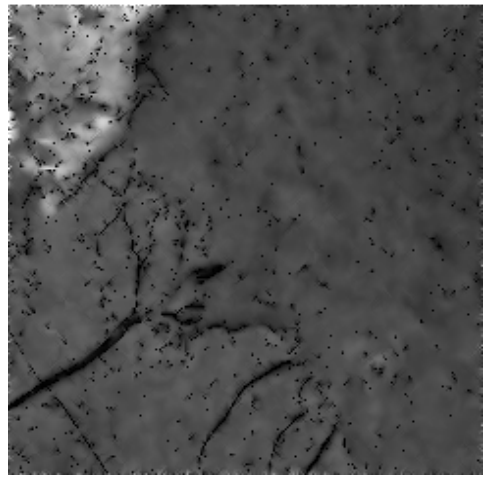

(d) Refined Lee

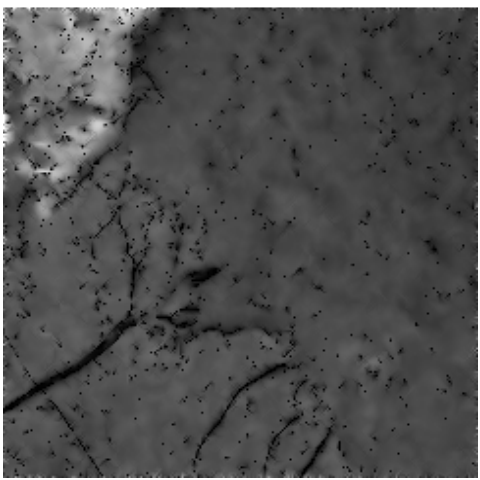

(e) Wiener

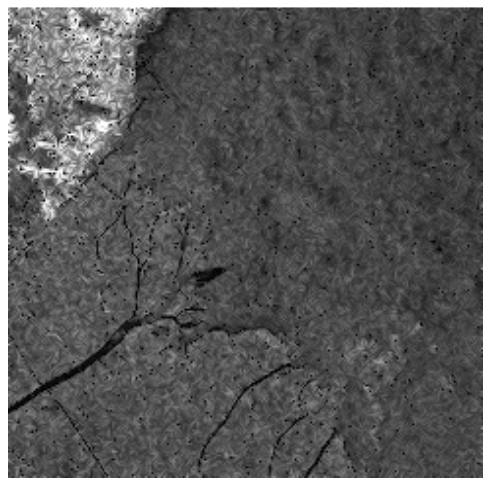

(f) NeuralShrink

Fig. 7: SAR with real speckled and filtered.

Table II. Assessment Parameters vs. Filters for Fig. 7.

| Filter | Assessment Parameters | | | | | |
|---|---|---|---|---|---|---|
| | MSD | NMV | NSD | ENL | DR | FOM |
| Original noisy image | - | 90.0890 | 43.9961 | 11.0934 | 2.5580e-017 | 0.3027 |
| En-Frost | 564.8346 | 87.3245 | 40.0094 | 16.3454 | 4.8543e-017 | 0.4213 |
| En-Lee | 532.0006 | 87.7465 | 40.4231 | 16.8675 | 4.4236e-017 | 0.4112 |
| Frost | 543.9347 | 87.6463 | 40.8645 | 16.5331 | 3.8645e-017 | 0.4213 |
| Lee | 585.8373 | 87.8474 | 40.7465 | 16.8465 | 3.8354e-017 | 0.4228 |
| Gamma-MAP | 532.9236 | 87.8444 | 40.6453 | 16.7346 | 3.9243e-017 | 0.4312 |
| Kuan | 542.7342 | 87.8221 | 40.8363 | 16.9623 | 3.2675e-017 | 0.4217 |
| Median | 614.7464 | 85.0890 | 42.5373 | 16.7464 | 2.5676e-017 | 0.4004 |
| Wiener | 564.8346 | 89.8475 | 40.3744 | 16.5252 | 3.2345e-017 | 0.4423 |
| DS | 564.8346 | 89.5353 | 40.0094 | 17.8378 | 8.5942e-017 | 0.4572 |
| EDS | 564.8346 | 89.3232 | 40.0094 | 17.4242 | 8.9868e-017 | 0.4573 |
| VisuShrink (HT) | 855.3030 | 88.4311 | 32.8688 | 39.0884 | 7.8610e-016 | 0.4519 |
| VisuShrink (ST) | 798.4422 | 88.7546 | 32.9812 | 38.9843 | 7.7354e-016 | 0.4522 |
| VisuShrink (SST) | 743.9543 | 88.4643 | 32.9991 | 37.9090 | 7.2653e-016 | 0.4521 |
| SureShrink | 716.6344 | 87.9920 | 32.8978 | 38.3025 | 2.4005e-015 | 0.4520 |
| NormalShrink | 732.2345 | 88.5233 | 33.3124 | 36.8464 | 6.7354e-016 | 0.4576 |
| BayesShrink | 724.0867 | 88.9992 | 36.8230 | 36.0987 | 1.0534e-015 | 0.4581 |
| Non-log SWT | 300.2841 | 86.3232 | 43.8271 | 11.2285 | 1.5783e-016 | 0.4577 |
| Non-log DWT | 341.3989 | 87.1112 | 39.4162 | 16.4850 | 1.0319e-015 | 0.4588 |
| NeuralShrink | 869.3422 | 91.8464 | 32.4231 | 39.2384 | 3.1423e-015 | 0.4601 |